\documentclass[5p,preprint]{elsarticle}

\usepackage{hyperref}
\usepackage{xcolor, soul}
\usepackage{nomencl}
\usepackage{framed}

\usepackage{multicol} 

\usepackage{nomencl} 

\usepackage{lineno}

\usepackage{graphicx}
\usepackage{subfigure}
\usepackage{amsmath,amssymb}
\usepackage{multirow}
\usepackage{algorithm}
\usepackage[]{algorithmic}
\usepackage{siunitx}

\usepackage[T1]{fontenc}
\usepackage[utf8]{inputenc}
\usepackage{soulutf8}
\usepackage[super]{nth}

\usepackage{color}
\usepackage{soul}

\usepackage[capitalise, noabbrev]{cleveref}

\begin{document}
	
	\begin{frontmatter}
		
		\title{Remaining useful life prediction of Lithium-ion batteries using spatio-temporal multimodal attention networks}
		\author[DFKI,TUKaiserslautern]{Sungho Suh\corref{mycorrespondingauthor}}
		\cortext[mycorrespondingauthor]{Corresponding author}
		\ead{Sungho.Suh@dfki.de}
        \author[TUKaiserslautern]{Dhruv Aditya Mittal}
		\author[DFKI,TUKaiserslautern]{Hymalai Bello}
        \author[DFKI,TUKaiserslautern]{Bo Zhou}
        \author[CRAN]{Mayank Shekhar Jha}
		\author[DFKI,TUKaiserslautern]{Paul Lukowicz}
		
		\address[DFKI]{German Research Center for Artificial Intelligence (DFKI), Kaiserslautern, Germany}
		\address[TUKaiserslautern]{Department of Computer Science, RPTU Kaiserslautern-Landau, Kaiserslautern, Germany}
		\address[CRAN]{Centre de Recherche en Automatique de Nancy (CRAN), University of Lorraine, Nancy, France}

		\begin{abstract}
            Lithium-ion batteries are widely used in various applications, including electric vehicles and renewable energy storage. The prediction of the remaining useful life (RUL) of batteries is crucial for ensuring reliable and efficient operation, as well as reducing maintenance costs. However, determining the life cycle of batteries in real-world scenarios is challenging, and existing methods have limitations in predicting the number of cycles iteratively. In addition, existing works often oversimplify the datasets, neglecting important features of the batteries such as temperature, internal resistance, and material type. To address these limitations, this paper proposes a two-stage RUL prediction scheme for Lithium-ion batteries using a spatio-temporal multimodal attention network (ST-MAN). The proposed ST-MAN is to capture the complex spatio-temporal dependencies in the battery data, including the features that are often neglected in existing works. Despite operating without prior knowledge of end-of-life (EOL) events, our method consistently achieves lower error rates, boasting mean absolute error (MAE) and mean square error (MSE) of 0.0275 and 0.0014, respectively, compared to existing convolutional neural networks (CNN) and long short-term memory (LSTM)-based methods. The proposed method has the potential to improve the reliability and efficiency of battery operations and is applicable in various industries.
		\end{abstract}
		
	\end{frontmatter}
	

\makenomenclature
\setlength{\nomitemsep}{-\parskip} 

\renewcommand*\nompreamble{\begin{multicols}{2}}
\renewcommand*\nompostamble{\end{multicols}}
\begin{table*}[!t]
  \begin{framed}
    \printnomenclature
  \end{framed}
\end{table*}
\nomenclature{ADLSTM}{Adaptive Dropout LSTM}
\nomenclature{Ah}{Ampere Hour}
\nomenclature{CNN}{Convolution Neural Network}
\nomenclature{EOL}{End of Life}
\nomenclature{ELM}{Extreme Learning Machine}
\nomenclature{FC}{Fully Connected}
\nomenclature{FLOPs}{Floating Point Operations}
\nomenclature{FPC}{First Prediction Cycle}
\nomenclature{GPR}{Gaussian Process Regression}
\nomenclature{GRU}{Gated Recurrent Units}
\nomenclature{HS}{Health State}
\nomenclature{HUST}{Huazhong University of Science and Technology}
\nomenclature{ISSA-MKELM}{Sparrow Search Algorithm}
\nomenclature{LSTM}{Long Short Term Memory}
\nomenclature{MCT}{Minimum Cycle Threshold}
\nomenclature{MAPE}{Mean Absolute Percentage Error}
\nomenclature{MSE}{Mean Squared Error}
\nomenclature{MKELM}{Multiple Kernel ELM}
\nomenclature{PSO}{Particle Swarm Optimization}
\nomenclature{RMSE}{Root Mean Squared Error}
\nomenclature{RUL}{Remaining Useful Life}
\nomenclature{SOC}{State of Charge}
\nomenclature{ST-MAN}{Spatio-Temporal Multimodal Attention Network}
\nomenclature{TCN}{Temporal Convolutional Network}

\section{Introduction}
\label{introduction}
        Lithium-ion batteries have become indispensable power sources across diverse applications, spanning from electric vehicles and renewable energy storage to consumer electronics and industrial systems \citep{dos2021lithium}. As their significance continues to grow, accurate prediction of the Remaining Useful Life (RUL) of these batteries assumes paramount importance. RUL prediction not only optimizes resource utilization but also enhances operational reliability and minimizes maintenance expenses, safeguarding the efficiency and longevity of battery-driven systems. In today's industrial landscape, reliable and efficient Lithium-ion battery operation underpins a spectrum of outcomes, from extended electric vehicle ranges to optimized renewable energy systems. Nonetheless, predicting RUL reliably is a formidable challenge due to the intricate interplay of factors governing battery degradation.

        Past attempts to predict RUL have yielded diverse methodologies encompassing model-based and data-driven approaches. Model-based methods use mathematical equations to build mechanism models to capture internal electrochemical reactions and prognosticate RUL. These models mainly consider internal attenuation mechanism factors, including loss of lithium and loss of active material in the electrodes \citep{birkl2017degradation, tian2021capacity}. Empirical models, such as exponential models and polynomial models, have been utilized to predict the degradation trend of Lithium-ion battery capacities \citep{zhang2019validation}. In addition, adaptive filter techniques, including the Kalman filter, have been adopted to update the model parameters \citep{zhang2022weight}. However, these approaches often exhibit significant RUL prediction errors when confronted with real-world complexities and uncertainties, despite the advantages of interpretability and the lack of a training procedure associated with data-driven methods.
        
        Conversely, data-driven methods have garnered substantial attention for their adaptability and ability to learn from data without necessitating exhaustive knowledge of underlying physics. Data-driven methods predict the RUL by training the degradation data of the Lithium-ion batteries with statistical and machine learning algorithms. Various machine learning algorithms, including support vector regression (SVR) \citep{zhao2018novel}, naive Bayes (NB) \citep{ng2014naive}, Gaussian process regression (GPR) \citep{zhang2020identifying}, convolutional neural networks (CNN) \citep{zhou2020state, ma2023two}, and long short-term memory (LSTM) networks \citep{li2019remaining, ma2019remaining, tong2021early}, have been utilized to predict the RUL of Lithium-ion batteries.
        However, data-driven methods require a substantial volume of historical degradation data for model training and exhibit limitations, notably in capturing intricate correlations between various degradation influencers. The majority of prevailing techniques predominantly focus on discharge capacity as a health indicator, failing to account for pivotal factors substantially shaping battery performance and degradation. In particular, existing data-driven methods require discharge capacity data from more than 25\% of total charge-discharge cycles, potentially overlooking sudden degradations in battery performance.

        Furthermore, conventional methods for RUL prediction grapple with limitations that curtail their efficacy in real-world scenarios \citep{salkind1999determination, miao2013remaining}. Conventional deep learning models, for instance, attempt to predict discharge values for later timestamps/cycles by inputting a percentage of the initial discharge capacity. Additionally, these data-driven approaches assume uniformity in data sources, presupposing that both training and test data originate from identical sensors under comparable operating conditions or share identical distributions.

        These conventional approaches face challenges in iteratively forecasting the number of cycles until battery end-of-life (EOL), relying on static models ill-suited for adapting to dynamic operational variations. Moreover, they tend to oversimplify datasets, disregarding crucial variables such as temperature, internal resistance, and material type \citep{chen2023overview, ma2023two}. Neglecting these essential parameters compromises the models' ability to accurately capture the intricacies of battery degradation mechanisms. In summary, the limitations of prevailing RUL prediction methods are twofold. Firstly, many struggle to dynamically forecast RUL across multiple cycles, limiting their applicability in dynamic real-world settings. Secondly, these approaches often overlook vital factors like temperature, internal resistance, and material type, resulting in imprecise and overly simplistic predictions.

        Furthermore, precise identification of a battery's health state (HS) is paramount for accurate RUL prediction. Unfortunately, many conventional hybrid methods fall short of recognizing the first prediction cycle (FPC), which signifies the commencement of the unhealthy stage \citep{suh2020supervised, suh2022generalized}. Neglecting to pinpoint the FPC can lead to suboptimal RUL predictions, particularly when the HS of the battery exhibits minimal differences in the run-to-failure training dataset.

        To overcome these limitations, in this paper, we propose a novel two-stage RUL prediction scheme for Lithium-ion batteries employing a spatio-temporal multimodal attention network (ST-MAN) architecture, aimed at addressing the critical challenge of RUL estimation in real-world scenarios where precise EOL information is often unavailable. 
        In the initial stage, our approach delineates the transition to the unhealthy state, effectively segmenting degradation data into discernible health stages. Subsequently, in the second stage, we predict the RUL as a percentage in the unhealthy state after the FPC.
        By considering a broader range of features, including discharge capacity, charge capacity, temperature, internal resistance, and charge time, we improve prediction accuracy and robustness. To improve the performance of the RUL prediction and generality of the model, we propose the ST-MAN which can adeptly capture intricate spatio-temporal connections inherent in multimodal battery degradation data. Our novel ST-MAN architecture for the RUL prediction combines CNN, LSTM, and spatio-temporal attention units to effectively capture intricate degradation patterns. Through a series of experiments on two public Lithium-ion battery datasets including the MIT dataset \citep{severson2019data} and HUST dataset \citep{ma2022real}, we demonstrate the effectiveness of the proposed two-stage scheme and the network architecture.
        
        The primary contributions of this paper can be summarized as follows.
        \begin{itemize}
            \item we introduce a two-stage RUL prediction scheme, involving the prediction of FPC in the first stage and the degradation pattern prediction in the second stage.
            \item we automatically annotate the degradation pattern as a percentage by determining the FPC to train the proposed ST-MAN model.
            \item we introduce an innovative ST-MAN architecture, combining CNN, spatio-temporal attention, and LSTM units, to capture intricate degradation patterns effectively.
            \item we validate our proposed method through rigorous experimentation, showcasing its superiority over existing CNN and LSTM-based methods on widely recognized battery degradation datasets.
        \end{itemize}
        
        The remainder of this paper is structured as follows. \cref{sec:relatedwork} introduces the related work. \cref{sec:method} provides the details of the proposed two-stage RUL prediction scheme and ST-MAN architecture. \cref{sec:experiments} presents the descriptions of the datasets used in our experiments and provides quantitative and qualitative experimental results on the two datasets. Finally, \cref{sec:conclusion} concludes the paper.


 \section{Related Work}
\label{sec:relatedwork}
        
        The landscape of data-driven models for predicting RUL in Lithium-ion batteries has witnessed significant development. Deep learning techniques have proven effective in enhancing RUL prediction, thereby augmenting predictive energy management capabilities \citep{wang2021critical}. Recurrent Neural Networks (RNNs) and Long Short-Term Memory (LSTM) algorithms have emerged as prominent strategies \citep{zhang2018long, hong2019synchronous, zhang2017lstm}. Dynamic LSTM variants, such as the work by \cite{song2018lithium}, have explored online RUL prediction employing indirect voltage measures for health index creation, deviating from traditional capacity-based approaches. Ensemble methods coupled with LSTM, as proposed by \cite{liu2019deep}, have accounted for uncertainties using Bayesian model averaging. \cite{ren2021method} integrated Particle Swarm Optimization (PSO) with LSTM, enabling efficient parameter estimation.
        
        LSTM derivatives like Gated Recurrent Units (GRUs) have also garnered attention. Existing works such as \cite{hannan2020state} addressed State of Charge (SOC) estimation under varying temperatures, while \cite{huang2019convolutional} introduced convolutional GRUs, and \cite{cui2020state} employed attention mechanisms within GRUs for battery prognostics. Complementary approaches include the application of Convolutional Neural Networks (CNNs), often combined with other techniques for enhanced performance. \cite{zhou2020state} incorporated causal and dilated convolutions to capture local contextual information. Hybrid architectures like CNNs combined with Bi-LSTM, as showcased by \cite{yang2020remaining}, have effectively integrated feature extraction and temporal modeling. Fusion strategies such as RNN-CNN combinations, as presented by \cite{zhao2020lithium}, have demonstrated merit in feature extraction and time dependency management.
        
        Extreme Learning Machine (ELM) has garnered attention due to its rapid learning speed and generalization performance, proving stable in RUL estimation \citep{wang2021critical, ma2020capacity, zhang2023indirect}. Variations like Kernel ELM (KELM) \citep{heidari2019efficient}, Multiple Kernel ELM (MKELM) \citep{liu2022remaining}, and the incorporation of MKELM with the Sparrow Search Algorithm (ISSA-MKELM) \citep{zhang2023indirect} have emerged to adapt to complex data relationships. Hybrid models have also surfaced; for instance, \cite{wang2022online} adopted a Bi-LSTM scheme with an attention mechanism for State of Health (SOH) estimation. Meanwhile, \cite{tong2021early} innovatively combined Adaptive Dropout LSTM (ADLSTM) and Monte Carlo simulation for RUL prediction, achieving accuracy comparable to methods utilizing more data.
        
        Incorporating domain knowledge, \cite{ma2023two} integrated CNN and Gaussian Process Regression (GPR) for multi-stage RUL prediction. Similarly, \cite{yao2022two} harnessed time series data and regression models for multi-cycle RUL prediction. Several works focus on using two stages for RUL prediction: \cite{xu2022novel} for smart manufacturing systems, \cite{song2021attention} for prognostic health management, \cite{qu2024remaining} for railroad systems, and \cite{zhang2022remaining} for mechanical equipment degradation. The data used to implement such methods differs from battery data, which is more complex due to varying physical elements and hidden chemical reactions inside the battery \citep{singh2021data, richardson2019battery, kouhestani2023data}. \cite{liu2019simultaneous} proposed an end-to-end intelligent fault diagnosis and RUL prediction framework on ball-bearing data, which contains features such as load ratings, bearing life, fatigue, failure modes, and vibration analysis. In contrast, the battery data contains features like charge/discharge cycles, energy density, power density, internal resistance, lifespan, failure modes, etc. Our work presents the application of fault diagnosis and RUL prediction on battery data where the physical and chemical properties vary from bearings data.
        
        Building upon our prior work \citep{mittal2023two}, which introduced a two-stage LSTM-based RUL prediction framework, this paper presents a novel RUL prediction method with the proposed ST-MAN architecture. By capturing intricate spatio-temporal dependencies, this method aims to address limitations in existing approaches and advance the accuracy and robustness of RUL prediction for Lithium-ion batteries.

	\section{Proposed Method}
        \label{sec:method}
        
        \begin{figure*}[!t]
        \centering
        \includegraphics[width=\linewidth]{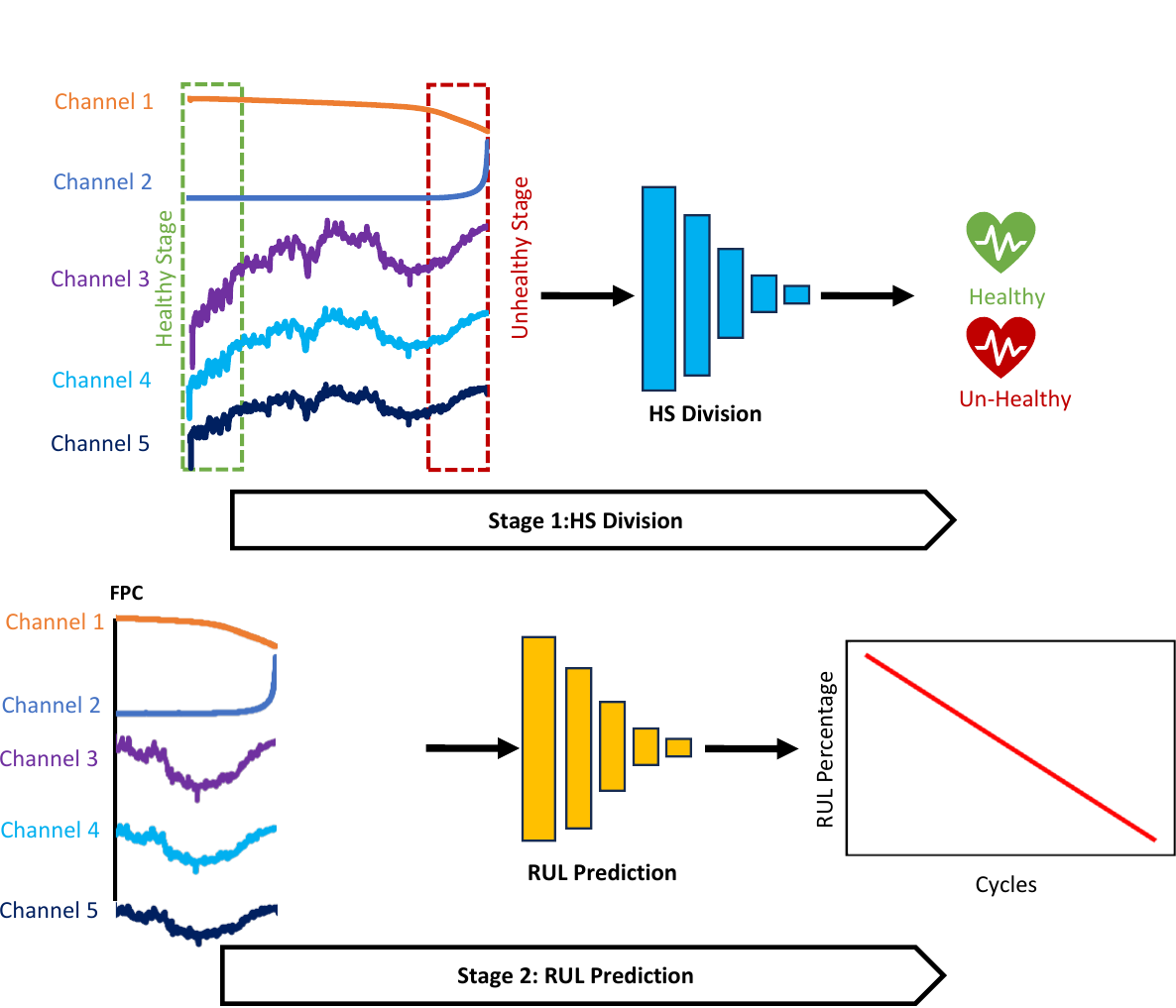}
        \caption{Overview of the proposed framework. Stage 1 is the health state (HS) division step using the LSTM classifier and Stage 2 is a remaining useful life (RUL) prediction step using the proposed spatio-temporal multimodal attention network (ST-MAN) architecture.}
        \label{proposed_framework}
        \end{figure*}

        In this section, we present our innovative RUL prediction framework, depicted in \cref{proposed_framework}, strategically crafted to address the intricacies of diverse degradation patterns observed in distinct battery cells. Departing from conventional methodologies, our framework adopts a two-stage paradigm. It initiates by categorizing the data from each cycle into either a healthy or unhealthy state, a pivotal classification step that markedly enhances the accuracy of RUL predictions. Our framework has two stages: an HS division stage and a RUL prediction stage. Within the HS division stage, our goal is to distinguish between healthy and unhealthy states in the degradation patterns of the battery cells. This discrimination plays a crucial role in identifying the FPC, signaling the commencement of the unhealthy phase—a pivotal factor for precise RUL predictions. 

        The FPC prediction is needed because typically, any system remains in its nominal state of functioning in the initial stages of operation, gradually leading to failure as time advances. This remains the case across the domains (fuel cells, rotary machine components like turbines, bearings, etc.) mainly because most of the degradation mechanisms have slow dynamics (crack propagation, battery discharge degradation, etc.). Moreover, the need for prognostics (i.e. prediction of end of life) is felt when the system is beyond the initial stages of operation i.e. after an initial 40-50\% of useful life has been consumed. 
        Usually, RUL forecasts are mainly needed for predictive maintenance and as such, their real utility becomes apparent when the real End of life (EOL) approaches. Given these two aspects, prognostics of most of the systems are generally not done from the very first second of operation and are usually triggered/” switched-on” after the system has functioned for some considerable part of the time (say 40-50\%). 
        For example, as proposed in the paper \citep{saxena2008metrics} thus, leading to such a practice since then within the community) the “True RUL” is considered constant (straight line) from the start till a certain point in time owing to the general understanding that assessing prognostics efficiency right from the beginning is not useful.
        
        Once the FPC is established, we define the subsequent data as degradation data. Utilizing an extensive array of features, including discharge capacity, charge capacity, charging time, temperature, and internal resistance, we forecast the RUL as a percentage from the FPC to the end-of-life (EOL). To effectively capture intricate degradation patterns, we propose the ST-MAN, which integrates CNN, LSTM, and spatio-temporal attention units.
                
        \subsection{First Prediction Cycle Decision}
        
        \begin{figure*}[!t]
        \includegraphics[width=\linewidth]{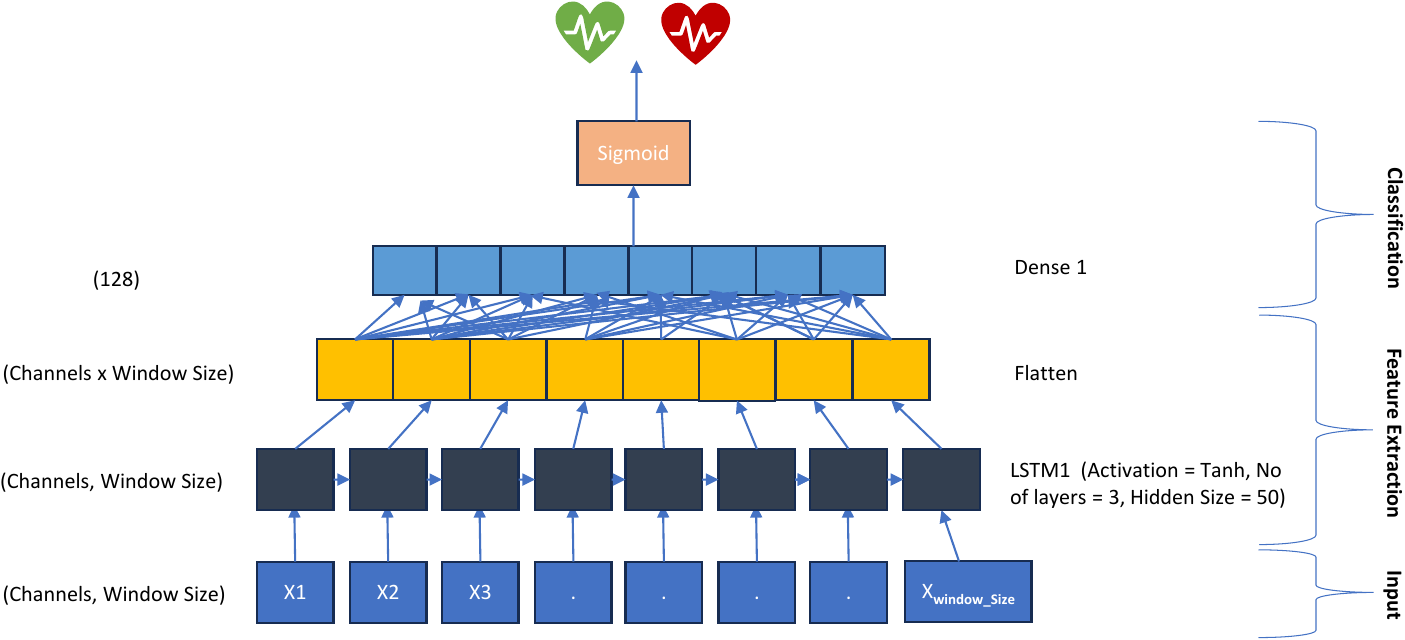}
        \label{fig:HS_classifier}
        \caption{HS division classifier network architecture based on the LSTM units.}
        \end{figure*}

        In this subsection, we address the intricate challenge of predicting time-series data with limited availability and inherently non-linear degradation patterns. Time-series data in the context of Lithium-ion battery health is notably influenced by a multitude of factors, including usage patterns, load variations, and dynamic operating conditions. These complexities hinder the precise measurement of the internal reactions occurring inside the battery. To mitigate the data limitations, we employ a sliding window approach, characterized by a window size of 50 and a step size of 1, including the different number of channels depending on the dataset. A window size of 50 is selected by referring to the works presented in \citep{fermin2020identification, attia2022knees, paulson2022feature, severson2019data, li2021one, saxena2022convolutional}

        The first stage of our proposed approach is centered on determining the FPC. This pivotal point signifies the transition from a healthy state to an unhealthy state within the battery cell degradation data. However, annotating the data to distinguish between healthy and unhealthy segments for training the HS division network poses a significant challenge. To overcome this limitation, we adopt a unique approach wherein we designate the initial 10\% of the cycles as the healthy state and the final 10\% as the unhealthy state in the training dataset, as illustrated in \cref{proposed_framework}.

        This strategy not only decreases the need for manual labeling but also enhances the accuracy of HS division, particularly in cases where ground truth HS labels are unavailable in most open-access Lithium-ion battery datasets.  
        
        The percentage value equal to 10\% is selected because the early life of the Lithium-ion battery is defined between the third cycle and 40\% of the total number of cycles of the battery and the end of life is defined after 80\% of the battery’s total number of cycles \citep{strange2023online}. Hence, we have used as training data for our model the first 10\% of the cycles labeled as healthy, and the last 10\% of the battery cycles labeled as unhealthy to guarantee the data is between the above-mentioned boundaries.
        
\begin{figure}[!t]
        \includegraphics[width=\linewidth]{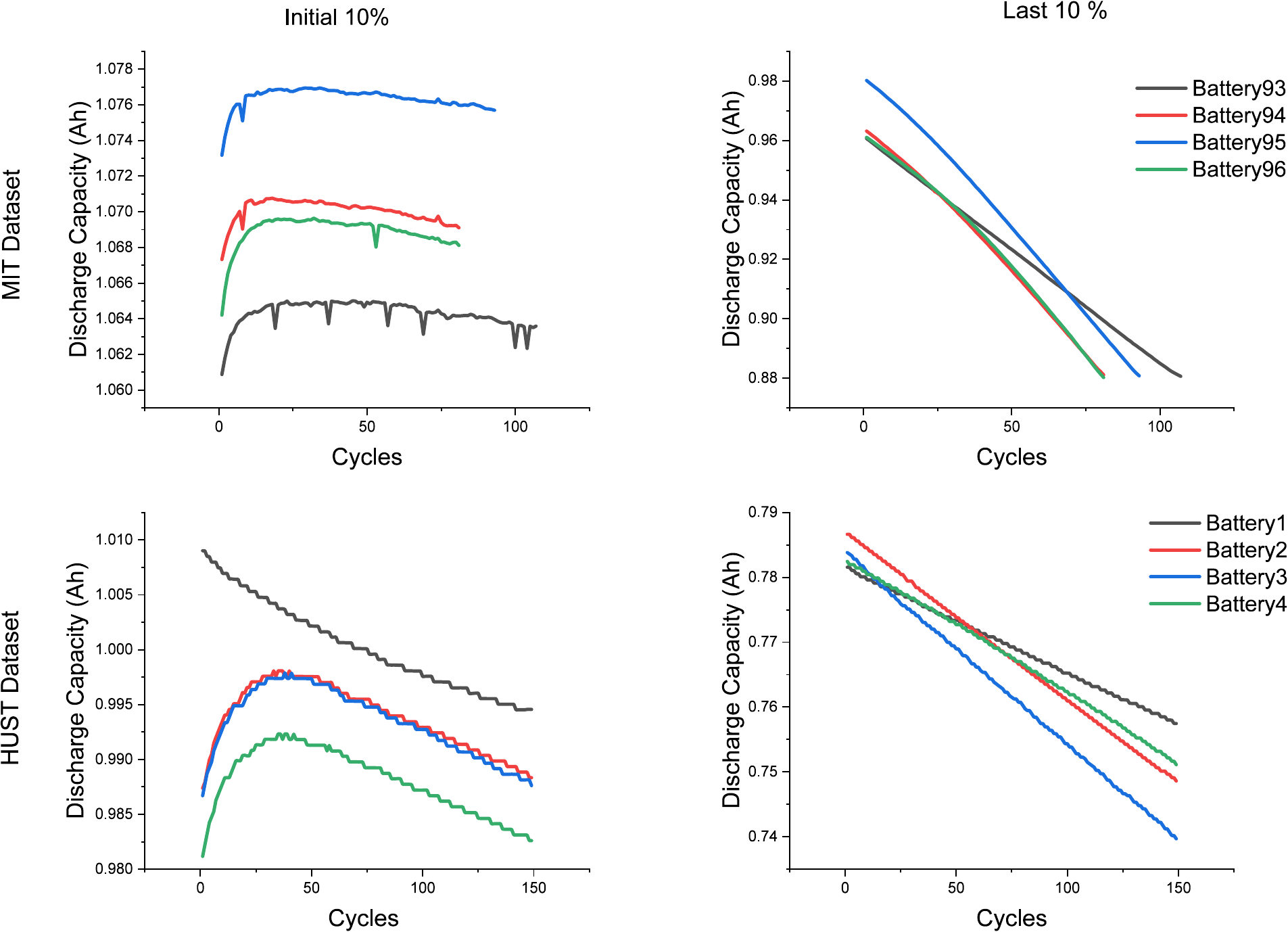}
        \label{fig:initial_and_last_both_datasets}
        \caption{Discharge Capacity at initial and last 10\% of the total cycles examples from MIT (Upper) and HUST dataset (Lower)}
        \end{figure}

        To provide further justification, \cref{tab:initial_last} demonstrates how the discharge capacity varies the initial and last 10\% of the total number of battery cycles, it is evident that the discharge capacity remains relatively stable during the initial 10\% of the data, whereas significant fluctuations are observed during the last 10\%.
        Moreover, the distribution of cycles in the initial and last 10\% of the total number of cycles along with the distribution of the total number of cycles in different batteries is presented in \cref{tab:cycles_inofmation}. To provide visualization, \cref{fig:initial_and_last_both_datasets} shows how the discharge capacity curves look during the initial and last 10\% of representative battery life in both datasets.

        \begin{table}[h]
            \caption{Mean standard deviation of discharge capacity during initial and last 10 \% of total cycles}
            \label{tab:initial_last}%
            \centering
            \begin{tabular}{lcc}
            \hline
            Dataset & Initial 10\%  & Last 10\% \\
            \hline
            MIT \citep{severson2019data}   & 0.002   & 0.024  \\
            HUST \citep{ma2022real}  & 0.004   & 0.014   \\
            \hline
        \end{tabular}
        \end{table}
            
            \begin{table}[h]
            \caption{Information related to number of cycles}
            \label{tab:cycles_inofmation}%
            \centering
            \begin{tabular}{lccc}
            \hline
            Dataset & All Data &  Initial 10\% & Last 10\% \\ 
            \hline
            MIT \citep{severson2019data} & 810 $\pm$ 369  & 75 $\pm$ 36& 80 $\pm$ 36 \\
            HUST \citep{ma2022real} & 1888 $\pm$ 387 & 183 $\pm$ 38 & 188 $\pm$ 38  \\
          \hline
        \end{tabular}
        \end{table}
    
        Consider $X^i=[x_1^i,...x_n^i]$ as input data, representing all features of the $i$-th battery cell. A sliding window of size $n_w$ is applied to accommodate the $n_f$ features, with $x_j^i \in \mathbb{R}^{n_f \times n_w}$ containing the battery cycle data within that window. The corresponding HS labels are defined as:

        \begin{equation}
            \label{eq:HSLabel}
            {y_{HS}}_j^i=\left\{
        	\begin{array}{@{}ll@{}}
        	0, & \text{if}\ j<\text{EOL}^i\times p \\
        	1, & \text{if}\ j>\text{EOL}^i\times (1-p)
        	\end{array}\right.
        \end{equation}
        where ${y_{HS}}_j^i$ denotes the HS label of $x^i_j$, EOL$^i$ represents the total number of cycles in the $i$-th battery cell data, and $p$ is the percentage of the total degradation process set to 10\%. Consequently, 20\% of the training data is labeled for training the HS division network model in the FPC decision stage. The shape of the variable $x^i_j$ is equal to [Window Size, Number of Channels]. For instance, the input dimension for training the model on MIT \citep{severson2019data} dataset is [50, 7], where 50 is window size and 7 is the number of features.

        The HS division model employs a 1D LSTM architecture, as illustrated in \cref{fig:HS_classifier}. This network model incorporates two LSTM modules, each comprising four layers. Following the LSTM layers, one fully connected (FC) layer with sigmoid activation functions is utilized to classify the HS. For training the HS division model, we use the binary cross entropy (BCE) loss function, quantifying the discrepancy between predictions and corresponding HS labels:
        \begin{equation}
            \label{eq:BCELoss}
            L_{BCE} = E_{{y_{HS}^i},{\hat{y}_{HS}}^i}[{y_{HS}^i}\log {\hat{y}_{HS}}^i + (1-{y_{HS}^i})\log (1-{\hat{y}_{HS}}^i)]
        \end{equation}
        where ${\hat{y}_{HS}}^i$ and ${y_{HS}^i}$ denote the HS division prediction and the corresponding HS label of the $i$-th battery cell data, respectively.

        After the HS division model is trained, it can categorize both the remaining training and test data into healthy and unhealthy states without the necessity for a predefined threshold value. By assimilating the feature distinctions between labeled healthy and unhealthy data during training, the model can discern degradation patterns across the entire dataset. Determining the FPC is crucial, marking the initial stages of Lithium-ion battery deterioration. However, identifying the FPC poses a challenge due to typically weaker degradation features during this phase compared to those in the labeled unhealthy state.

        To identify the relevant FPC, we employ a straightforward continuous trigger mechanism. In this setup, the battery cell is classified as unhealthy when consecutively predicted as such by the HS division model for a defined number of instances. This trigger mechanism helps mitigate unnecessary oscillations and uncertainties in FPC determination, being a widely adopted strategy in HS division approaches \citep{lei2018machinery, suh2020supervised}. In our study, the FPC is determined when the HS division model designates the input data as unhealthy for five consecutive instances, contributing to a more reliable indication of the FPC. To further improve the robustness of FPC determination, we incorporate additional measures within the continuous trigger mechanism. The pseudo-code for determining the FPC point is mentioned in \cref{alg:fpc_prediction_algo}.

        In particular, we introduce a temporal element to the trigger mechanism to ensure that the FPC is not identified too early in the battery's life cycle. This added temporal constraint guards against prematurely labeling the battery as unhealthy, which could result in inaccurate RUL predictions.
        
        Under this extended mechanism, the battery is considered to have reached its FPC only if the HS division model classifies it as unhealthy for five consecutive times and this classification occurs beyond a defined threshold cycle, which we refer to as the minimum cycle threshold (MCT). We selected the threshold value for consecutive prediction of the unhealthy stage as five to eliminate oscillations between healthy and unhealthy stage predictions and to pinpoint the FPC accurately. When calculating the FPC point for each battery, we observed that in most cases, no oscillation was present. However, to safeguard our model from potential random oscillations. The MCT is a dynamic value calculated as a percentage of the total cycle count. By setting an MCT, we ensure that the FPC identification occurs after a certain proportion of the battery's expected lifetime has elapsed, adding a temporal context to the determination process.
        
        The introduction of this temporal constraint is crucial for several reasons. First, it aligns with the practical reality that the FPC often occurs after a certain number of cycles, rather than immediately. Second, it prevents premature FPC identification, which can be especially beneficial in cases where initial battery cycles exhibit varying degrees of degradation before stabilizing. Finally, this temporal constraint enhances the adaptability of our framework to different battery types and operating conditions, as it accounts for variations in degradation behavior.

        \begin{algorithm}[!t]
            \small

            \caption{FPC Point Prediction Algorithm}
            \label{alg:fpc_prediction_algo}
            \textbf{Input:} Battery data $D$ \\
            \textbf{Output:} FPC point prediction $p$
            
            \begin{algorithmic}[1]
                \STATE \textbf{Step 1: Data Preparation}
                \STATE Divide $D$ into training set $D_{\text{train}}$ and test set $D_{\text{test}}$
                \FOR{each battery $b$ in $D$}
                    \STATE Annotate fixed windows of size 50 from the initial 10\% of $b$ as \text{Healthy}
                    \STATE Annotate fixed windows of size 50 from the last 10\% of $b$ as \text{Unhealthy}
                \ENDFOR
            
                \STATE \textbf{Step 2: Train the HS Division Model}
                
                \STATE \textbf{Step 3: FPC Prediction}
                \STATE \textbf{Input:} Use the unannotated data (Remaining 80\% from each battery in $D_{\text{train}}$ and 100\% for each battery in $D_{\text{test}}$)
                \STATE $threshold\_count \gets 5$
            
                \STATE $FPC\_points \gets []$
                \FOR{each battery $b$ in $D$}
                    \STATE $unhealthy\_count \gets 0$
                    \FOR{each fixed window $w$ in $b$}
                        \STATE $window\_count \gets window\_count + 1$
                        \STATE $HS \gets HS\_Division\_Model(w)$
                        \IF{$HS == \text{Unhealthy}$}
                            \STATE $unhealthy\_count \gets unhealthy\_count + 1$
                        \ELSE
                            \STATE $unhealthy\_count \gets 0$
                        \ENDIF
                        \IF{$unhealthy\_count = threshold\_count$}
                            \STATE $FPC\_points \gets FPC\_points + [window\_count - threshold\_count]$
                            \STATE \textbf{break}
                        \ENDIF
                    \ENDFOR
                \ENDFOR
                \RETURN $FPC\_points$
                
            \end{algorithmic}
        \end{algorithm}

        \subsection{Remaining Useful Life Prediction}
        In the stage of predicting RUL, our objective is to estimate the RUL percentage after the FPC. For this purpose, we introduce the ST-MAN architecture, a novel design adept at capturing intricate degradation patterns while maintaining computational efficiency. This architectural choice seeks a fine balance between effectiveness and computational resources, presenting a lightweight yet powerful model for deep learning in RUL prediction.


        To effectively train our RUL prediction model, our approach utilizes time-series data by applying sliding windows from the FPC to the EOL as inputs, while the RUL percentage serves as the ground truth for training. Since the precise RUL information is generally unavailable in real-world scenarios, we calculate the RUL percentage label after the FPC as follows.
        
        \begin{equation}
            \label{eq:RULLabels}
            {y_{L}}_j^i = \frac{\text{EOL}^i - j}{\text{EOL}^i - \text{FPC}^i}
        \end{equation}
        where ${y_{L}}_j^i$ represents the RUL label of $x_j^i$ post FPC, while EOL$^i$ and FPC$^i$ denote the EOL and FPC of the $i$-th battery cell, respectively. This formulation allows us to express the RUL as a percentage, providing a more practical and interpretable measure of battery health.
        
        Our approach fundamentally differs from conventional approaches that rely exclusively on discharge capacity as a health indicator and predict discharge capacity values as RUL estimates. 
        Recognizing the tendency of discharge capacity values to exhibit sudden drops, making accurate RUL predictions challenging, we define RUL as a linear percentage.
        This approach considers the period from FPC to EOL, enabling the prediction of remaining life cycles from the testing cycle, given its linear relationship with the period from FPC to the cycle point and the remaining life cycles from the cycle point to EOL.

        \begin{figure*}[!t]
        \includegraphics[width=\linewidth]{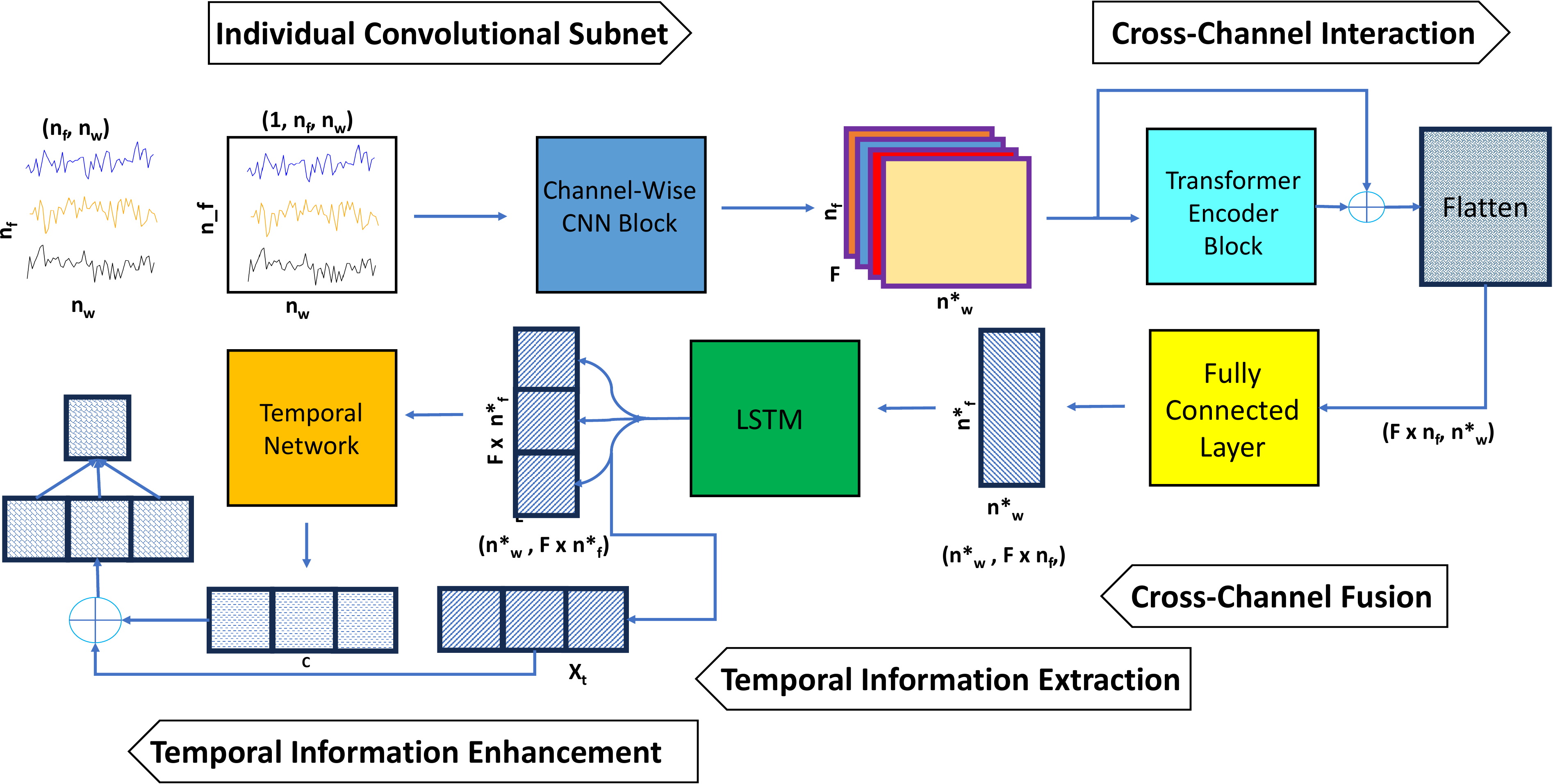}
        \caption{The architecture overview of the proposed spatial-temporal multimodal attention networks (ST-MAN). The network consists of convolutional operations, a cross-channel Transformer encoder block, a fully connected layer to fuse cross-channel information, an LSTM module to extract temporal information, and a temporal attention module to improve the temporal information extraction.}
        \label{fig:TransformerLSTM}
        \end{figure*}

\begin{table*}[!t]
\caption{\label{table:STMAN_model_architecture}
The details of the proposed ST-MAN network architecture for training on example MIT dataset \citep{severson2019data}}
\centering
\renewcommand{\arraystretch}{1.3}
\resizebox{\linewidth}{!}{
\begin{tabular}{ | c|c|c|c|c| } 
\hline
Network & Layers & Description &Activation & Output \\ 
\hline
\multirow{8}{5em}{Individual Convolution Sublet }
    & Input (7x50)   &                                       &  -     & - \\ 
    & Reshape & - & - & 1x7x50 \\ 
    & Permute & - & - & 1x50x7 \\ 
    
    &Conv2d    &  CNN (input = 1, output = 20, kernel size = (5,1), stride = (1, 1) )   &ReLU  & 20x46x7\\ 
    &Conv2d    &  CNN (input = 20, output = 20, kernel size = (5,1), stride = (2, 1) )              &ReLU  & 20x 21x 7 \\
    &Conv2d    &  CNN (input = 20, output = 20, kernel size = (5,1), stride = (1, 1) )              &ReLU  & 20x17x7\\
    &Conv2d    &  CNN (input =20, output = 20, kernel size = (5,1), stride = (2, 1) )              &ReLU  &20x7x7 \\
\hline
Channel Interaction & SelfAttention   &   Query(20,20), Key(20,20), Value(20,20)          &   Sigmoid  &  7x20x7 \\
\hline
\multirow{2}{5em}{Channel Fusion}
 & FC      &   Linear(140,40)        & - &  128x7x40  \\
 & Dropout      &  Dropout(0.1)       &  - &  128x7x40 \\
\hline
Temporal Fusion & LSTM      &  LSTM( input = 40, hidden size = 40)    &  Tanh &  7x40 \\
\hline
\multirow{2}{5em}{Temporal Fusion}
& FC      &  Linear(40,40)          &  Tanh &  40\\
& FC      &  Linear(40,1)         &  Softmax &  1\\
\hline
Prediction & FC      &  Linear(40,1)      &  - &  1 \\
\hline

\end{tabular}}
\end{table*}

        The proposed ST-MAN architecture is depicted in \cref{fig:TransformerLSTM}, along with its detailed layer-wise description in \cref{table:STMAN_model_architecture}. This architecture is specifically tailored to the unique characteristics of our battery degradation data, which is inherently multimodal and temporally dependent.
        We start by applying convolutional operations to the input data $x_j^i \in \mathbb{R}^{n_f \times n_w}$, which represents battery degradation data with different features $n_f$ using a sliding window of length $n_w$. Importantly, we treat each channel individually during this process to emphasize local context and account for the distinct contributions of each channel.
        To understand the relationships between different channels and determine the relative importance of each channel, we employ a transformer encoder block. This block enables the model to learn interactions between channels and their significance in the overall degradation process. The relative importance calculated by the transformer is then added back into the previous input data.
        
        After learning channel interactions, we proceed to fuse the channel information using an FC layer. Unlike the self-attention mechanism, the FC layer allows different features within the same sensor channel to have varying weights, enhancing the feature fusion and capturing the unique contributions of each.
        To model temporal dependencies across the sequence of data, we utilize LSTM units. These units enable the network to understand the global temporal relationships between different time steps within the input window.
        Considering that not all time steps contribute equally to RUL prediction, it is vital to assess the importance of features at each time step in the sequence. To achieve this, we generate a global contextual representation through a weighted average sum of hidden states at each time step. These weights are determined by a temporal self-attention layer. Since the feature at the last time step represents the entire sequence, the resulting global representation is reintegrated into the previous feature. We introduce a trainable multiplier parameter for the global representation, affording the model flexibility in deciding whether to utilize or disregard the generated global representation. Finally, an FC layer is applied to the represented feature to obtain the final RUL output.

        To train the RUL prediction model effectively, we define the RUL prediction loss, which measures the difference between the predicting RUL and the corresponding RUL labels as a percentage.
        \begin{equation}
            \label{eq:RULloss}  
            L_{RUL}	 = L_{MAE} + L_{RMSE} + L_{MAPE}
        \end{equation}
        \begin{equation}
            \label{eq:MAEloss}
            L_{MAE} = \frac{1}{\text{EOL}^i - \text{FPC}^i}\sum_{j=\text{FPC}^i}^{\text{EOL}^i} |{y_{L}}_j^i - {\hat{y_L}}_j^i |
        \end{equation}
        \begin{equation}
            \label{eq:RMSEloss}
            L_{RMSE} = \sqrt{\frac{1}{\text{EOL}^i - \text{FPC}^i}\sum_{j=\text{FPC}^i}^{\text{EOL}^i} ({y_{L}}_j^i - {\hat{y_L}}_j^i )^2}
        \end{equation}
        \begin{equation}
            \label{eq:MAPEloss}
            L_{MAPE} = \frac{1}{\text{EOL}^i - \text{FPC}^i}\sum_{j=\text{FPC}^i}^{\text{EOL}^i} \frac{\arrowvert {y_{L}}_j^i - {\hat{y_L}}_j^i \arrowvert}{ {y_{L}}_j^i }
        \end{equation}
        where ${y_{L}}_j^i$ and ${\hat{y_L}}_j^i$ represent the corresponding RUL label and RUL prediction of the $i$-th battery cell data, respectively. $L_{MAE}$ denotes the mean absolute error (MAE) between the predicting RUL and the corresponding RUL labels, $L_{RMSE}$ is root mean square error (RMSE), $L_{MAPE}$ is mean absolute percentage error (MAPE), and the final RUL prediction loss is the sum of the MAE, RMSE, and MAPE. 
        This comprehensive loss function ensures that the model learns to make precise predictions while considering both the magnitude and relative importance of prediction errors. Combining MAE, RMSE, and MAPE into a composite loss function offers a comprehensive evaluation of prediction accuracy, considering error magnitude, relative errors, and sensitivity to outliers. This approach ensures a balanced assessment, robustness to individual metric idiosyncrasies, and simplicity in interpretation, providing a holistic evaluation of model performance across various error dimensions. By incorporating these architectural elements and the corresponding loss functions, our proposed ST-MAN architecture is equipped to effectively capture and learn from the complex degradation patterns exhibited by Lithium-ion batteries, leading to accurate and robust RUL predictions.
	
\section{Experimental Results}
\label{sec:experiments}

        \subsection{Datasets}
        
        \begin{figure}[!t]
            \subfigure[]{
            \includegraphics[width=0.45\columnwidth]{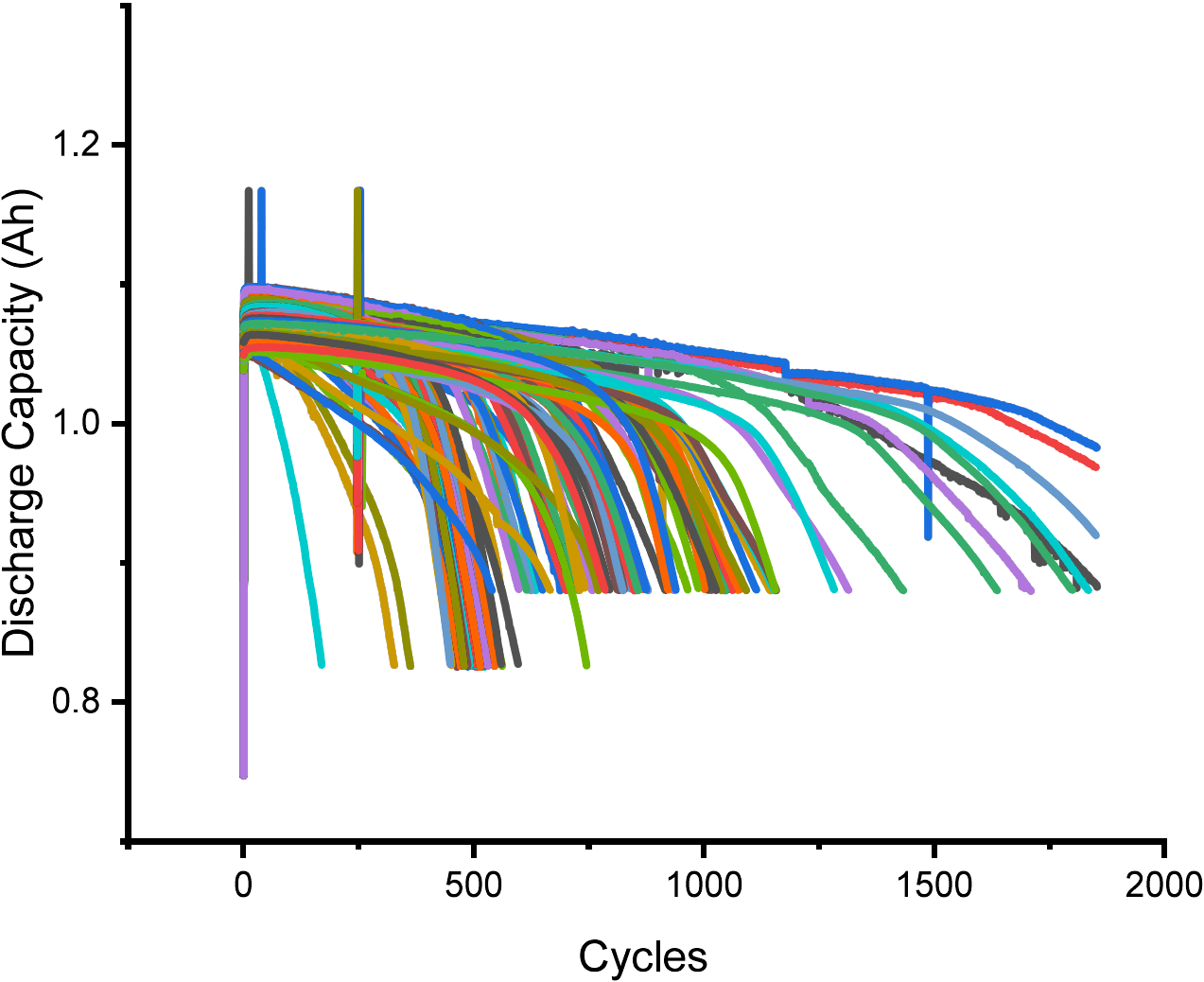}
            }
            \subfigure[]{
            \includegraphics[width=0.45\columnwidth]{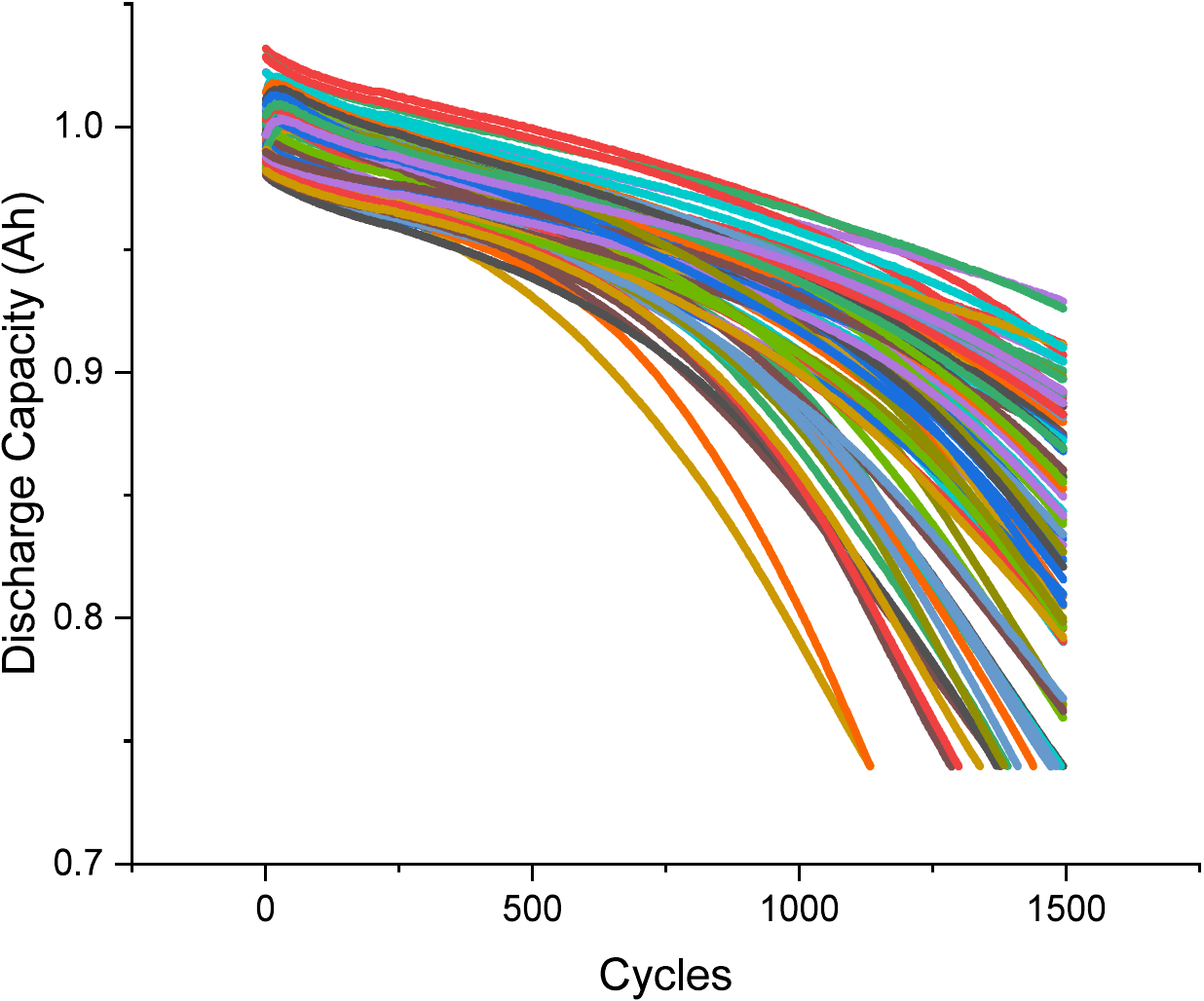}
            }
            \caption{Discharge capacity degradation patterns of (a) 124 Lithium-ion battery cells from the MIT dataset \citep{severson2019data}, (b) 77 Lithium-ion battery cells from HUST dataset \citep{ma2022real}}
            \label{fig:Dataset}
        \end{figure}
        
        To evaluate the proposed framework on the real battery cell degradation data, we used two different datasets: the MIT dataset \citep{severson2019data} and the HUST dataset \citep{ma2022real}. While the MIT dataset is widely recognized in the field, the HUST dataset represents one of the latest additions, offering valuable battery discharge data. The MIT dataset is the product of a collaborative effort between the Toyota Research Institute, Stanford, and MIT \citep{severson2019data}. This dataset comprises 124 commercial lithium-ion battery cells, specifically the A123 Systems model APR18650M1A, boasting a nominal capacity of 1.1 Ah. These batteries were subjected to rigorous cycling under fast-charging conditions. The experimentation involved cycling the batteries within horizontal cylindrical fixtures using a 48-channel Arbin LBT potentiostat, all within a controlled forced convection temperature chamber maintained at $30^\circ C$. The key specifications of the cells in this dataset include a nominal capacity of 1.1 Ah and a nominal voltage of 3.3 V. Charging was executed following either a one-step or two-step fast-charging policy, denoted as 'C1(Q1)-C2'. Here, C1 and C2 represent the initial and subsequent constant-current steps, while Q1 signifies the state-of-charge (SOC, \%) at which the current transitions between the two steps. Following this, charging at 1C CC-CV ensued, concluding at 80\% SOC, which is EOL. The upper and lower cutoff potentials adhere to the manufacturer's specifications at 3.6 V and 2.0 V, respectively. The dataset spans a wide range of cycle lives, varying from 150 to 2,300 cycles, with an average of approximately 810 cycles and a standard deviation of 340. Each cycle corresponds to a complete battery discharge and recharge. This rich dataset comprises crucial parameters, including cycle number, internal resistance, temperature statistics (minimum, average, maximum), and charge and discharge capacities. In \cref{fig:Dataset} (a), we present a visualization of the discharge capacities of the 124 cells, demonstrating the significant variability in cycle numbers among the lithium-ion batteries.

        The HUST dataset \citep{ma2022real} is provided by the Huazhong University of Science and Technology and involves 77 commercial batteries subjected to over 140,000 charge-discharge cycles. These batteries, all LFP/graphite A123 APR18650M1A with a nominal capacity of 1.1 Ah and a nominal voltage of 3.3 V, undergo testing with various multi-stage discharge protocols. However, they follow an identical fast-charging protocol in two thermostatic chambers at $30^\circ C$. The number of cycles ranges from 1,100 to 2,700 cycles, with an average of 1,898 and a standard deviation of 387, each representing information about discharge capacity, charge capacity, and charge voltage. \cite{ma2022real} claim that this is the largest dataset for diverse protocols. Moreover, we used the feature generation mechanism used in \citep{ma2022real} to generate two additional features, where the difference of the charge voltage and charge capacity curve between each cycle and the \nth{10} cycle is taken into account, making a total of $5$ features for each cycle. The high performance of differences in charge capacity features and voltage capacity features are inspired by \cite{severson2019data} and \cite{jiang2021bayesian}, respectively. As illustrated in \cref{fig:Dataset} (b), the discharge capacity for all 77 cells in the HUST dataset exhibits diverse patterns. Notably, the distinct properties of both datasets make them valuable resources for conducting experiments and analyzing results, thereby contributing to the development of robust solutions for RUL prediction tasks.

    \subsection{Implementation Details and Evaluation Metrics}

        \begin{figure}[!t]
            \centering
            \includegraphics[width=\columnwidth]{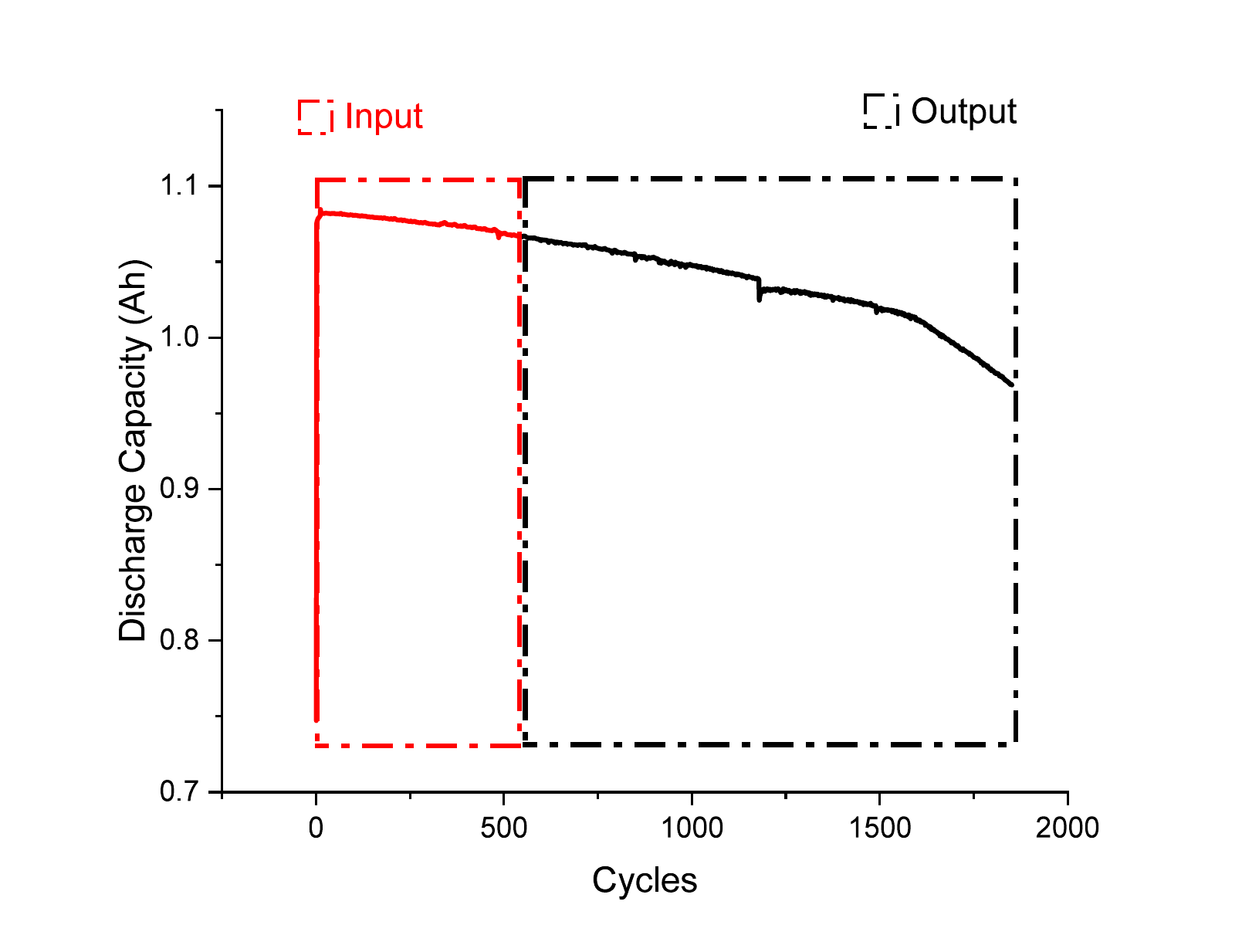}
            \caption{The conventional RUL prediction scheme with input and output on the same battery cell}
            \label{conventional}
        \end{figure}

        The experiments were carried out using Python scripts within the PyTorch framework. The models were optimized using the Adam optimizer with a learning rate of 0.0001, $\beta_1$ at 0.9, and $\beta_2$ at 0.99. The training process involved a batch size of 8, running for 100 epochs with early stopping, implemented with a patience of 20 epochs to mitigate overfitting.

        To demonstrate the advantage of our proposed framework, we conducted comparisons with established models, specifically the Temporal Convolutional Network (TCN) \citep{zhou2020state} and ADLSTM \citep{tong2021early}. These conventional methods traditionally rely on discharge capacity as the sole input and presume prior knowledge of the EOL. The RUL prediction scheme for conventional methods, as illustrated in \cref{conventional}, employs 40\% of the battery cell data as input, predicting the remaining 60\% as output. To address variations in cycle numbers across datasets, padding was implemented to align input and output sizes. For example, on the MIT dataset, the input and output sizes were set at 894 and 1342, respectively, while for the HUST dataset, sizes of 1072 and 1608 were utilized.

        Additionally, we compare the proposed ST-MAN with the TCN and ADLSTM, all using our two-stage RUL prediction scheme and the same number of input features. These experiments highlight the effectiveness of the ST-MAN architecture for RUL prediction.

        In the initial stage of our framework, the goal is to predict the HS of the battery cell at various cycle points and identify the FPC. In our experiments, we labeled the first 10\% of the total cycle data for each cell as the healthy state and the last 10\% as the unhealthy state, leaving the remaining 80\% for FPC prediction. To mitigate overfitting, we performed 5-fold cross-validation. For instance, on the MIT dataset, 99-100 battery cell data were used for the training dataset, and 24-25 battery cell data for the test dataset, iteratively changing the training and test dataset split to include all cells in the test dataset.  To pass input of uniform size we use the fixed sliding window of size 50 to train the HS classification model and the RUL prediction model. For training the HS classification model in Stage 1 we show the number of cycles labeled as healthy (initial 10\%) and unhealthy (last 10\%) in \cref{tab:cycles_inofmation}.

        \begin{figure*}[!t]
        \centering
        \includegraphics[width=\linewidth]{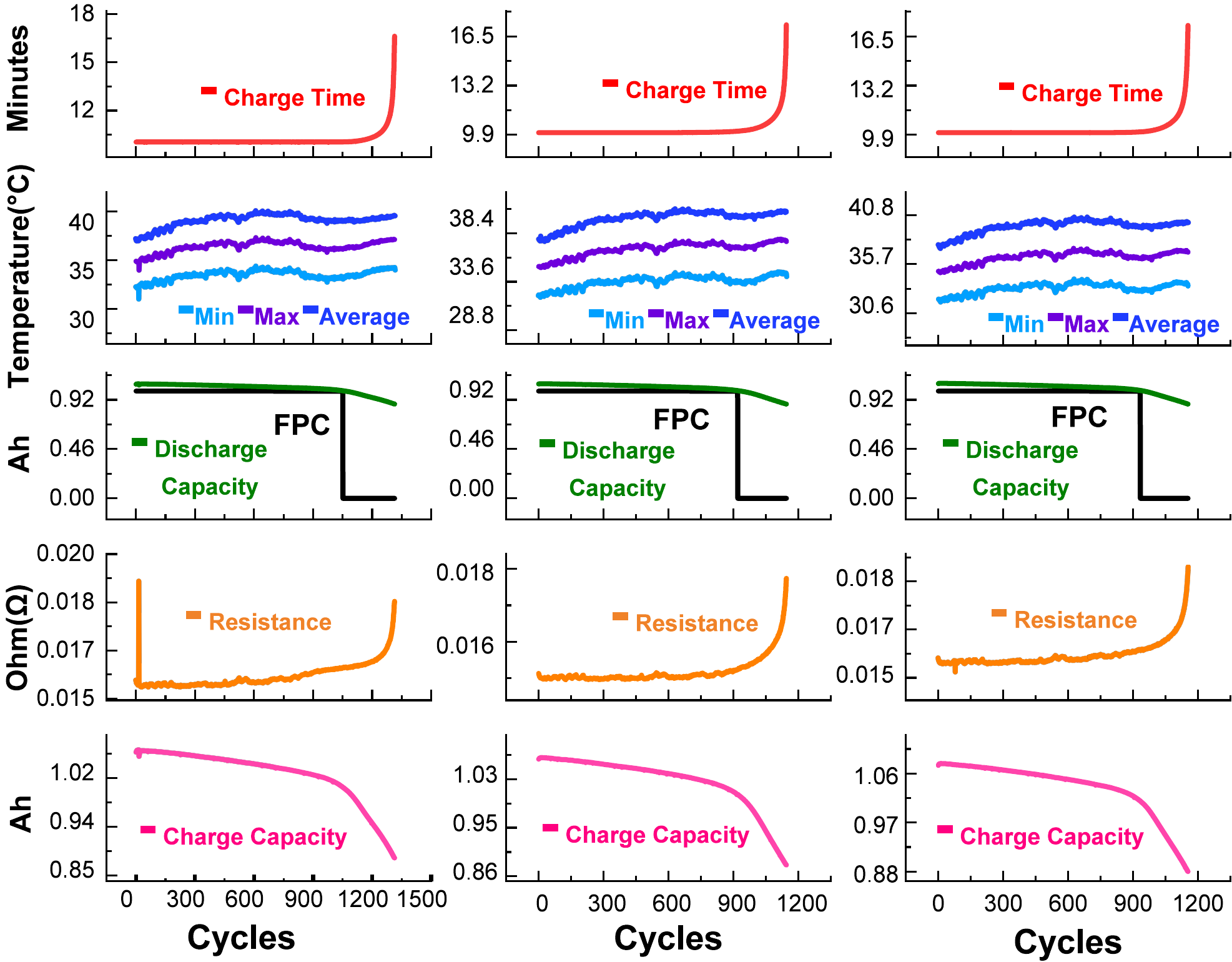}
        \caption{Illustration of first predicted cycle (FPC) decisions based on seven features for three battery cells from the MIT dataset: Discharge capacity, internal resistance, charge capacity, average temperature, minimum and maximum temperature, and charge time. The cells depicted are numbered \textbf{First} 101, \textbf{Second} 102, \textbf{Third} 103.}
        \label{FPC_Testing}
        \end{figure*}
        
        Moreover, we use 7 and 5 input features from the MIT and HUST datasets, respectively. The input features of the MIT dataset include discharge capacity, internal resistance, charge capacity, average temperature, minimum and maximum temperature, and charge time. The HUST dataset input features comprise charge voltage, discharge capacity, charge capacity, the difference of the charge voltage between each cycle and the 10th cycle ($\Delta V_{i-10}=V_i - V_{10}$), and the difference of the charge capacity between each cycle and the 10th cycle ($\Delta Q_{i-10}=Q_i - Q_{10}$). T
        
        For the evaluation and comparison of the proposed framework with conventional methods, three evaluation metrics—MAE, MSE, and MAPE—were employed, defined as follows:
        
        \begin{equation}
            \label{eq:MAE}
            MAE = \frac{1}{N} \sum_{i=1}^{N}|y_i-\hat{y}_i| 
        \end{equation}
        
        \begin{equation}
            \label{eq:MSE}
            MSE = \frac{1}{N} \sum_{i=1}^{N}(y_i-\hat{y}_i)^{2}
        \end{equation}
        
        \begin{equation}
            \label{eq:MAPE}
            MAPE = \frac{100\%}{N} \sum_{i=0}^{N - 1} \frac{\arrowvert y_i - \hat{y}_i \arrowvert}{y_i}.
        \end{equation}
        where ${\hat{y}_i}$ represents the predicted RUL percentage and $y_i$ represents the ground truth. These evaluation metrics allow us to quantitatively assess the accuracy and performance of our models in predicting remaining useful life.

        \subsection{Results and Discussion}
        
\begin{figure}[!t]
        \includegraphics[width=\linewidth]{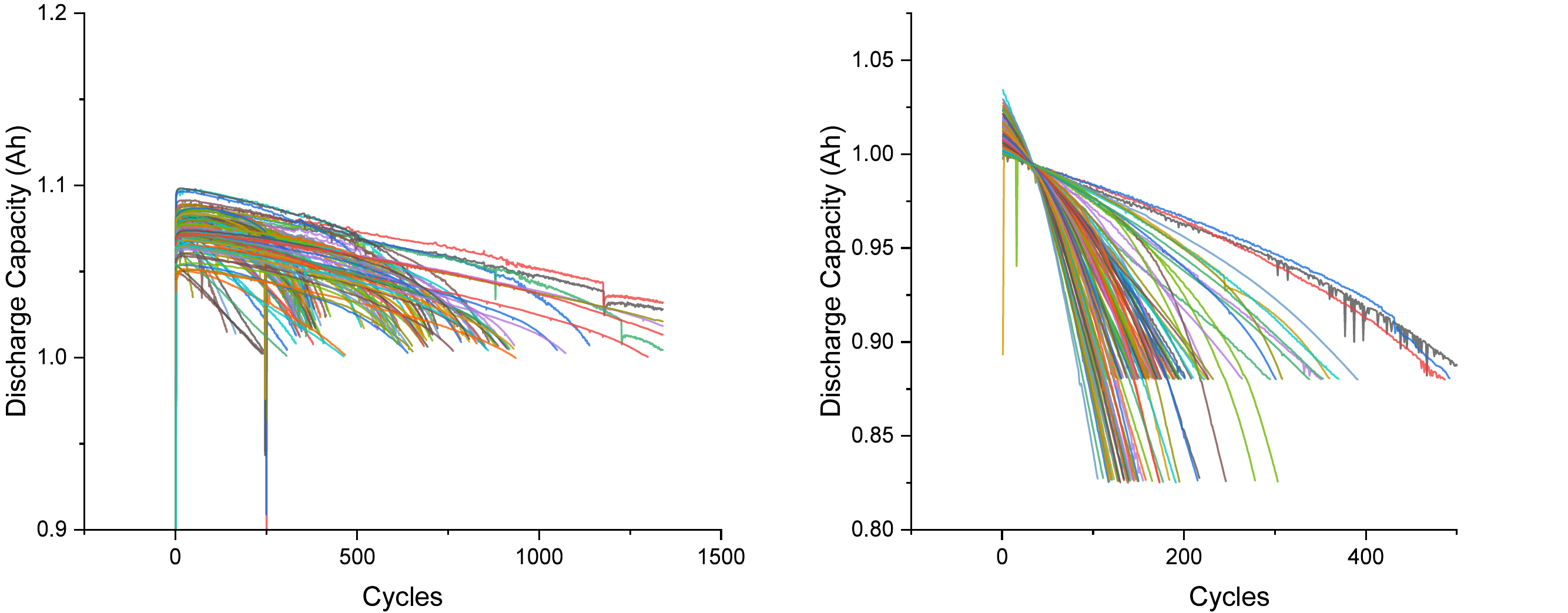}
        
        \caption{Discharge Capacities for MIT \citep{severson2019data} dataset before and after the FPC.}
        \label{fig:curves_before_and_after_FPC}
        \end{figure}

        In \cref{FPC_Testing}, instances of the health state (HS) division outcomes from the initial stage of our proposed scheme are demonstrated using the MIT dataset. The results illustrate a distinct classification between the healthy and unhealthy states. Significantly, the identification of FPCs occurred at precise junctures corresponding to the onset of degradation in both discharge and charge capacity values.
        To provide further insight into the FPC determination process, we include the percentage of discharge capacity at the FPC point for both datasets in \cref{tab:FPC}. This percentage represents the proportion of remaining discharge capacity when the FPC occurs.  For the MIT dataset, on average, the FPCs were identified at approximately 94\% of the total discharge capacity, while for the HUST dataset, they occurred at roughly 90\% of the total discharge capacity. This consistent trend across both datasets demonstrates the robustness and stability of our proposed first-stage approach in pinpointing the FPCs accurately. These findings reinforce the reliability of our method in identifying the critical transition points within battery degradation cycles, which is paramount for precise remaining useful life prediction. Furthermore, the discharge capacity curves before and after the FPC point in the MIT \citep{severson2019data} dataset can be visualized in \cref{fig:curves_before_and_after_FPC}, where the discharge capacity is relatively stable before the FPC point as compared to the discharge capacity after the FPC point. 

                  \begin{figure*}[!t]
        \includegraphics[width=\linewidth]{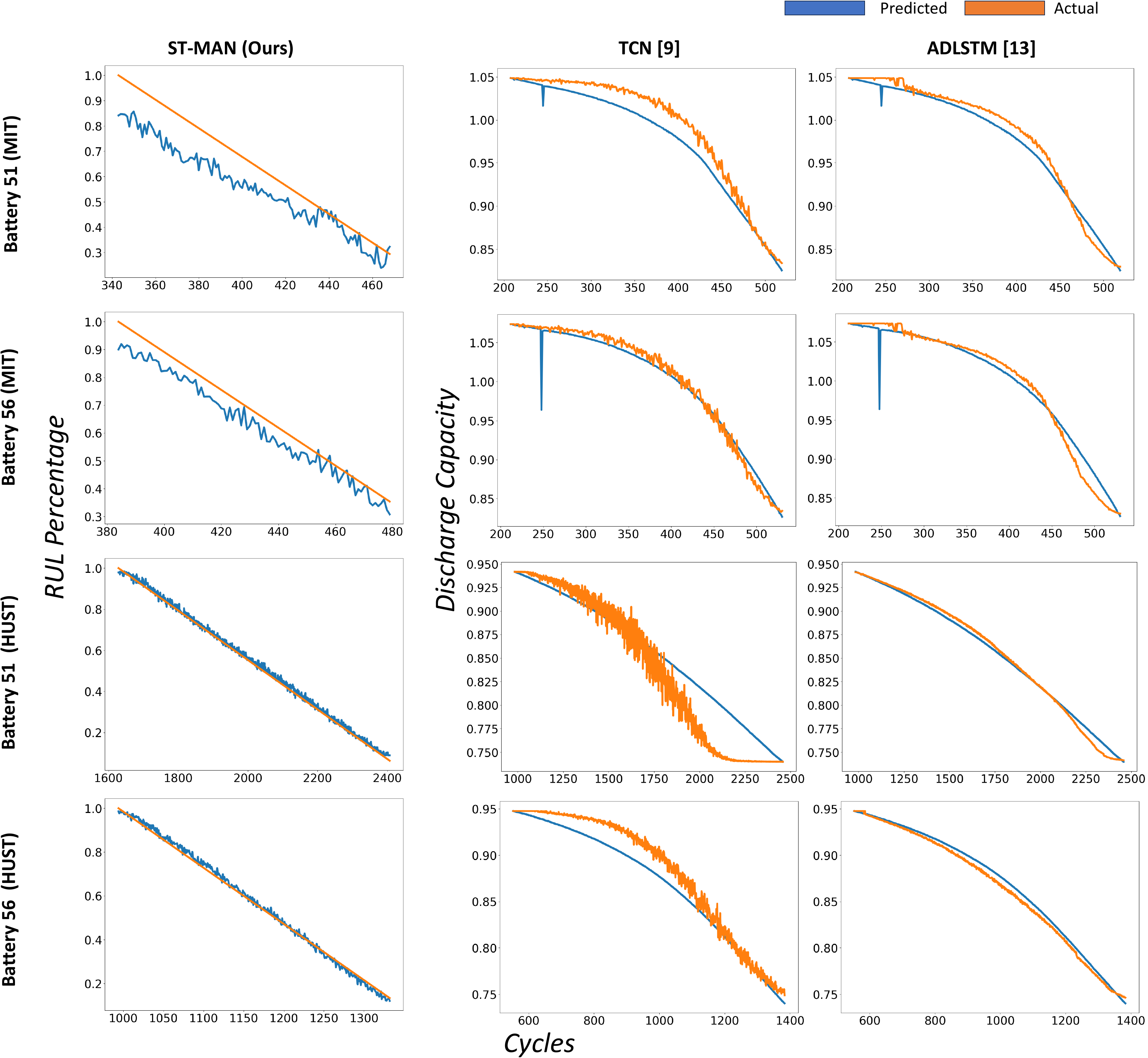}
        \caption{RUL percentage prediction, First two rows show RUL prediction on batteries 51 and 56 of MIT dataset, last two rows show RUL prediction on batteries 51 and 56 on HUST dataset}
        \label{fig:RUL_prediction_HUST_MIT}
        \end{figure*}
        
        \begin{table}[t]
        \caption{Percentage of Discharge Capacity at FPC}
        \label{tab:FPC}%
        \centering
        \begin{tabular}{lcc}
        \hline
        Dataset &  Train  &  Test\\
        \hline
        MIT  \citep{severson2019data} & 94.70   & 94.34   \\
        HUST  \citep{ma2022real}  & 90.771   & 86.69  \\
        \hline
        \end{tabular}
        \end{table}

        \begin{figure*}[!t]
        \includegraphics[width=\linewidth]{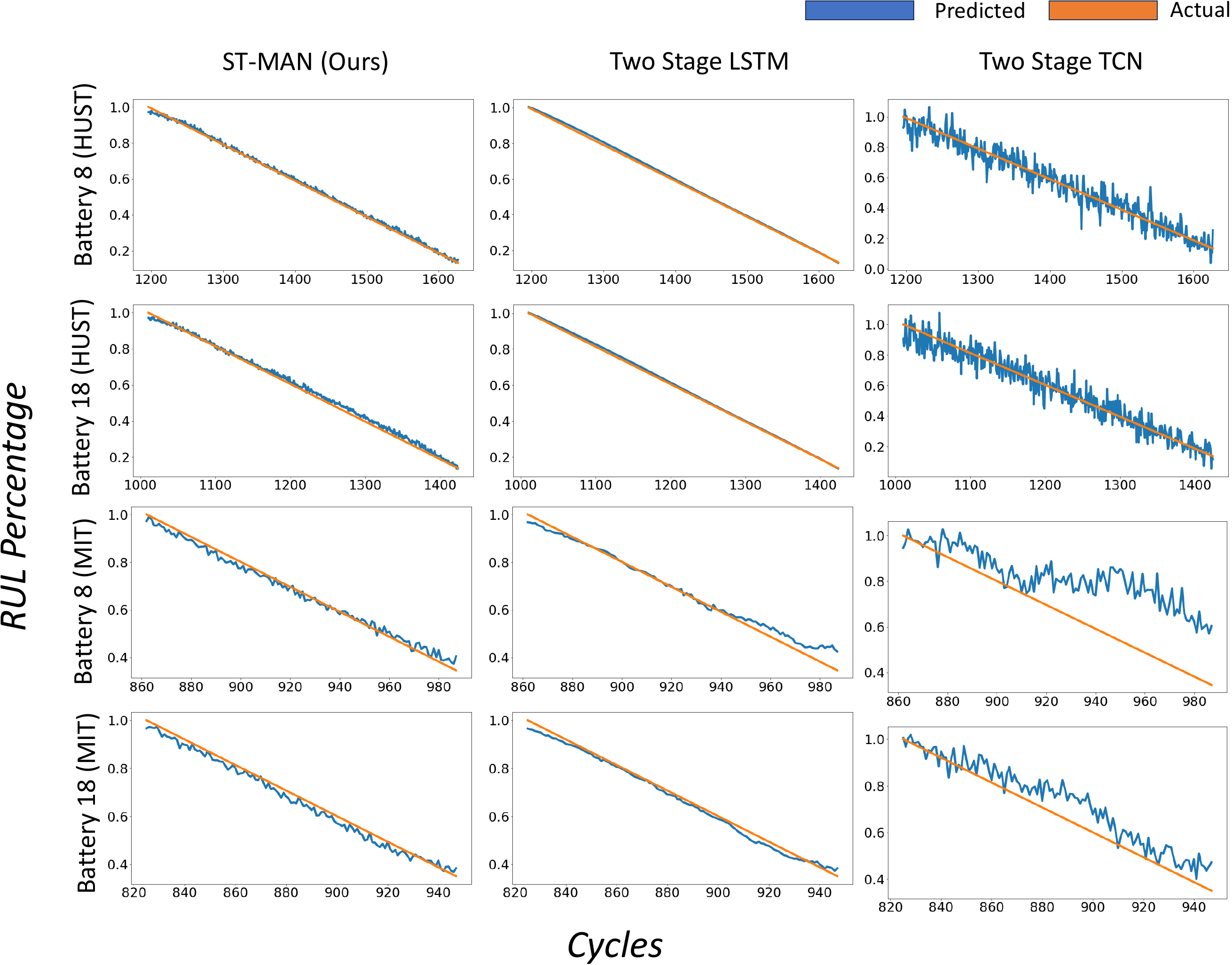}
        \caption{RUL percentage prediction, First two rows shows RUL prediction on batteries 8 and 18 of HUST dataset, last two rows show RUL prediction on batteries 8 and 18 on MIT dataset}
        \label{fig:RUL_prediction_HUST_MIT2}
        \end{figure*}

        \begin{table*}[!t]
        \caption{Evaluation of RUL prediction on the MIT and HUST datasets.}
        \label{tab:comparison}
        \centering
        \renewcommand{\arraystretch}{1.5}
        \resizebox{\linewidth}{!}{
        \begin{tabular}{cccc|ccc}
        \hline%
        \multirow{2}{*}{Methods} &  \multicolumn{3}{c}{MIT \citep{severson2019data}} & \multicolumn{3}{c}{HUST \citep{ma2022real}}\\
        \cline{2-7}
                &   MAE     &   MSE     & MAPE  & MAE   &MSE    & MAPE  \\
        \hline
        TCN \citep{zhou2020state}   & 0.0451 $\pm$ 0.0091    & 0.0065 $\pm$ 0.0046 & 0.1308 $\pm$ 0.0360  & 0.0226 $\pm$ 0.0081 & 0.0014 $\pm$ 0.0015 & 0.0776 $\pm$ 0.0237\\
        ADLSTM \citep{tong2021early} & 0.0461 $\pm$ 0.0146   & 0.0072 $\pm$ 0.0067 &0.1149 $\pm$ 0.0477 & 0.0237 $\pm$ 0.0030  &0.0006 $\pm$ 0.0002 &0.0834 $\pm$ 0.0148
        \\
        ST-MAN (Ours)               & \textbf{0.0275 $\pm$ 0.0046}   & \textbf{0.0014 $\pm$ 0.0005} & \textbf{0.0494 $\pm$ 0.0100}  & \textbf{0.0168 $\pm$	0.0091} & \textbf{0.0005 $\pm$ 0.0006} & \textbf{0.0384 $\pm$ 0.0193}\\
        \hline
        Two-stage TCN               & 0.1034 $\pm$ 0.0110   & 0.0160 $\pm$ 0.0027 & 0.1462 $\pm$ 0.0172  & 0.0939 $\pm$ 0.0389 & 0.0139 $\pm$ 0.0093 & 0.1670 $\pm$ 0.0821\\
        Two-stage LSTM              & 0.0293 $\pm$ 0.0043   & 0.0015 $\pm$ 0.0004 & 0.0544 $\pm$ 0.0077  & 0.0181 $\pm$ 0.0023 & 0.0006 $\pm$ 0.0001 & 0.0438 $\pm$ 0.0048\\
        \hline
        \end{tabular}
        }
        \end{table*}

        In \cref{fig:RUL_prediction_HUST_MIT}, we present a visual comparison of the RUL prediction results achieved by our proposed two-stage framework and conventional schemes \citep{zhou2020state, tong2021early}. It is worth noting that there is a fundamental difference in how RUL percentage prediction commences between the proposed method and conventional approaches, as depicted in \cref{fig:RUL_prediction_HUST_MIT}. Specifically, our proposed method initiates RUL percentage prediction after the FPC point, which signifies the starting point of battery degradation. Conversely, conventional methods base their predictions on data covering the first 40\% of the overall cycle, making the initial prediction points distinctly dissimilar.
        
        This disparity in prediction starting points has profound implications for the accuracy of RUL predictions. In the case of conventional methods, which necessitate prior knowledge of the EOL, the predicted RUL exhibits noticeable gaps when compared to the ground truth. Moreover, conventional models struggle to accurately anticipate sudden drops in RUL, often leading to discontinuous predictions. In contrast, our proposed method excels in producing smooth and stable RUL predictions. By commencing RUL estimation after the FPC point, our approach circumvents the need for prior knowledge of the EOL, resulting in more reliable and continuous predictions. This is particularly beneficial in real-world scenarios where exact EOL information may not be readily available.
        
        For a comprehensive evaluation of the RUL prediction models, we provide a detailed comparison in \cref{tab:comparison}, utilizing three prominent evaluation metrics. Notably, our proposed method consistently outperforms the two conventional methods across all these metrics. This superior performance underscores the effectiveness of our approach in accurately predicting RUL, all while obviating the need for prior knowledge of the EOL. These results highlight the potential of our two-stage framework in enhancing the precision of RUL prediction in various real-world applications.
        
        Furthermore, we compare the proposed ST-MAN with the TCN and LSTM, all employing our two-stage RUL prediction scheme and the same number of input features. \cref{fig:RUL_prediction_HUST_MIT2} demonstrates examples of the RUL prediction results on the MIT and HUST datasets by the proposed ST-MAN, two-stage TCN, and two-stage LSTM. Compared to the results obtained by conventional schemes in \cref{fig:RUL_prediction_HUST_MIT}, the proposed scheme improved the performance of the RUL prediction by the TCN and LSTM models. In \cref{tab:comparison}, the proposed ST-MAN also outperformed the two-stage TCN and two-stage LSTM. These experiments demonstrate the effectiveness of the ST-MAN architecture for RUL prediction and its potential to outperform conventional and alternative deep learning models in this critical task.
	
        \begin{table}[!t]
        \caption{Comparison of computation complexity and the number of parameters}
        \label{tab:computation}%
        \centering
        \begin{tabular}{@{}lll@{}}
        \hline
        Model & FLOPs  &  Parameters \\
        \hline
        TCN \citep{zhou2020state}    & 527,424   & 46,049  \\
        ADLSTM \citep{tong2021early}  & 1,003,648   &190,721   \\
        ST-MAN (Ours)   & 1,200,869   & 27,861   \\
        \hline
        \end{tabular}
        
        \end{table}
        
        In \cref{tab:computation}, we present a comprehensive overview of the computational complexities and performance metrics associated with the proposed method compared to the TCN and LSTM models when employing the two-stage RUL prediction scheme on the MIT dataset. Our assessment of computational costs encompasses both the number of network parameters and floating-point operations (FLOPs), a metric widely used to represent the execution time and overall computational complexity of neural network models \citep{jouppi2017datacenter}.
        
        Remarkably, as illustrated in \cref{tab:computation}, our proposed network architecture shows the smallest number of parameters. However, it is noteworthy that the FLOPs associated with the proposed ST-MAN model are relatively larger compared to the TCN and LSTM models. This signifies that our ST-MAN model exhibits a higher computational demand during execution. Nevertheless, the smaller number of weight parameters, despite the increased FLOPs, underscores the model's efficiency and demonstrates that it can achieve robust predictive performance with a leaner set of learnable parameters. This balance between FLOPs and parameters demonstrates the capability of the model to deliver favorable results while maintaining computational efficiency.
        
        \subsection{Ablation Study}
        In this section we discuss the ablation study performed to select the best set of parameters for our proposed model. \cref{tab:ablation_HS_Classifier} investigates the number of parameters required to train the HS Division Classifier for FPC prediction using various hyperparameters. Based on this analysis, we have chosen our HS classifier to consist of 1 LSTM block with 3 hidden layers inside it. This configuration strikes a balance between early FPC detection and the number of parameters, optimizing the performance of our model.
        
        \cref{tab:ablation_STMAN_architecture} shows how the RUL prediction performance is impacted after removing certain blocks from the proposed architecture. It can be easily observed that removing any of the subsections in the proposed model provided inferior performance in the performance metrics compared to the complete.

        \begin{table*}[!t]
    \caption{Ablation Study for HS Division Model by changing the number layers and the number of LSTM blocks in the model}
    \label{tab:ablation_HS_Classifier}
    \centering
    \resizebox{\linewidth}{!}{
    \begin{tabular}{c|ccc|ccc|ccc|ccc}
        
        \hline
        \multirow{3}{*}{Hidden Layers in each block} & \multicolumn{12}{c}{No of LSTM Blocks} \\
        \cline{2-13}
        & \multicolumn{6}{c|}{1} & \multicolumn{6}{c}{2} \\
        
        \cline{2-13}
        
        & \multicolumn{3}{c|}{MIT} & \multicolumn{3}{c|}{HUST} & \multicolumn{3}{c|}{MIT} & \multicolumn{3}{c}{HUST} \\\hline

        & Params & Train & Test & Params & Train & Test & Params & Train & Test & Params & Train & Test \\
        \cline{2-13}
        
        1 & 65,457 & 94.6 & 93.5  &52,657 & 90.1 & 83.9 & 85,857 & 94.5& 93.4 & 73108 & 92.2 & 85.6 \\
        2 & 85,857  &94.8 & 94.1 & 73,057&92.3 & 85.7& 126,657 & 96.4& 96.6 & 113908 & 92.4& 85.8\\
        3 & \textbf{106,257} & \textbf{96.8} & \textbf{98.4} & \textbf{93,457}& \textbf{91.8}& \textbf{85.7} & 167,457 & 96.2 &  98.6& 154708 & 91.4& 85.7\\
        4 & 126,657 & 96.8& 98.42  & 113,857& 92.4 &  85.8& 208,257 &95.4 &95.8 & 195508 & 91.3 & 86.8\\
        \hline
    \end{tabular}}
\end{table*}

        

        \begin{table*}[!t]
        \caption{Ablation study of ST-MAN architecture}
        \label{tab:ablation_STMAN_architecture}
        \centering
        \renewcommand{\arraystretch}{1.5}
        \resizebox{\linewidth}{!}{
        \begin{tabular}{cccc|ccc}
        \hline%
        \multirow{2}{*}{Methods} &  \multicolumn{3}{c}{MIT \citep{severson2019data}} & \multicolumn{3}{c}{HUST \citep{ma2022real}}\\
        \cline{2-7}
                &   MAE     &   MSE     & MAPE  & MAE   &MSE    & MAPE  \\
        \hline
        Without Initial Convnet  &  0.1646 $\pm$ 0.0115    &  0.0396 $\pm$ 0.0060 &   0.3180 $\pm$ 0.02691    & 0.2255 $\pm$ 0.0235 &  0.0727 $\pm$  0.0191 & 0.4329 $\pm$ 0.02857 \\
        
        Without Channel Interaction & 0.0500 $\pm$  0.0147  & 0.0046 $\pm$  0.0046 &0.1043 $\pm$ 0.0314 & 0.0471 $\pm$  0.0511 &0.0074 $\pm$ 0.0132& 0.0954 $\pm$ 0.1123
        \\
        Without Temporal LSTM  & 0.0358 $\pm$ 0.00542   & 0.0024 $\pm$ 0.0008 & 0.0728 $\pm$ 0.0116 &  0.0176 $\pm$  0.0042 & 0.0005 $\pm$ 0.0002   & 0.0331 $\pm$ 0.0072
        \\
        ST-MAN (Ours)               & \textbf{0.0275 $\pm$ 0.0046}   & \textbf{0.0014 $\pm$ 0.0005} & \textbf{0.0494 $\pm$ 0.0100}  & \textbf{0.0168 $\pm$ 0.0091} & \textbf{0.0005 $\pm$ 0.0006} & \textbf{0.0384 $\pm$ 0.0193}\\
        \hline
        
        \end{tabular}
        }
        \end{table*}

        \section{Conclusion}
        \label{sec:conclusion}
        In this study, we proposed a novel two-stage framework for RUL prediction of Lithium-ion batteries, addressing the critical challenge of RUL estimation in real-world scenarios where precise EOL information is often unavailable. Our proposed framework, which combines health state classification and RUL prediction, offers several key contributions and advantages. First, we introduced the concept of the FPC, which serves as a pivotal point to initiate RUL prediction. By identifying the FPC through health state classification, we effectively circumvent the need for prior knowledge of the EOL. This feature is particularly valuable in practical applications, where obtaining exact EOL information can be challenging or impossible.
        To achieve accurate health state classification and RUL prediction, we designed the ST-MAN, a novel architecture tailored to the unique characteristics of battery degradation data. ST-MAN contains CNNs, Transformer encoders, and LSTM units to capture intricate degradation patterns while maintaining computational efficiency. The use of multi-attention mechanisms and temporal self-attention enables ST-MAN to learn both local and global dependencies in the data, resulting in robust RUL predictions.
        Our experimental results demonstrate the effectiveness of the proposed framework and ST-MAN architecture. We compared our approach with conventional methods and alternative deep learning models, consistently achieving superior RUL prediction accuracy across multiple evaluation metrics. Notably, our method produces smooth and stable RUL predictions, addressing the challenges of discontinuity and inaccuracy associated with conventional methods.

        Furthermore, for the integration of an embedded solution, it is necessary to refine our strategy with a focus on reducing FLOPs in future iterations. In this work, we propose a hierarchical method comprising two stages. In the first stage, the healthy and unhealthy stages of the battery are determined. Subsequently, the second stage is activated when the battery status is confirmed as unhealthy. Both stages utilize high-performance LSTM-based models, which are computationally expensive. To mitigate the computational burden, it is recommended to revise the model architecture for the FPC detection (First stage) by transitioning to a CNN-based model, for example, employing knowledge distillation techniques. In the case of battery monitoring in embedded devices, only the first stage of our algorithm will be deployed. This decision is based on two main considerations: 1. When the battery of the embedded device is identified as unhealthy, the degradation of the device performance makes it an unreliable source for subsequent calculations, hence the second stage must be performed on an external device, such as the cloud. 2. By deploying only the first stage of our algorithm, the computational burden for the embedded device will be considerably alleviated, thereby improving its operational efficiency and resource utilization.
        
        Future work may involve further refinement of the framework, exploration of additional datasets, and deployment in real-world battery management systems. By advancing the field of battery health management and extending the lifespan of energy storage technologies, our research has the potential to revolutionize various industries, including the automotive and renewable energy sectors.

\section*{Acknowledgements}
This work was supported by Carl-Zeiss Stiftung under the Sustainable Embedded AI project (P2021-02-009) and by the European Union under the HumanE AI Network (H2020-ICT-2019-3 \#952026).

	\bibliography{ref}

\end{document}